\documentclass[letterpaper, 10 pt, conference]{ieeeconf}  

\IEEEoverridecommandlockouts                              
\usepackage[utf8]{inputenc}
\usepackage[T1]{fontenc}
\usepackage{graphicx} 
\usepackage[table,xcdraw]{xcolor}

\title{\LARGE \bf
A Hybrid Natural Language Generation System Integrating Rules and Deep Learning Algorithms
}

\author{Wei Wei$^{1}$, Bei Zhou$^{2}$, Georgios Leontidis$^{1}$\\\\
\parbox{3 in}
{\centering $^{1}$Department of Computing Science,\\
         University of Aberdeen\\
         \tt\small w.wei.19@abdn.ac.uk\\
         \tt\small georgios.leontidis@abdn.ac.uk}
         \hspace*{ 0.5 in}
         \parbox{3 in}
{ \centering $^{2}$School of Computer Science,\\ The University of Sydney\\
         \tt\small bei.zhou@sydney.edu.au}
}

\begin{document}

\maketitle
\thispagestyle{empty}
\pagestyle{empty}

\begin{abstract}

This paper proposes an enhanced natural language generation system combining the merits of both rule-based approaches and modern deep learning algorithms, boosting its performance to the extent where the generated textual content is capable of exhibiting agile human-writing styles and the content logic of which is highly controllable. We also come up with a novel approach called HMCU to measure the performance of the natural language processing comprehensively and precisely. 

\end{abstract}

\section{INTRODUCTION}

Nowadays, mainstream natural language generation (NLG) techniques fall into two categories, i.e. conventional rule-based approaches and deep learning algorithm-based approaches, each of which carries exclusive pros and cons. The former approaches can produce high-quality text with controllable context logic but suffering from the fact that the generated text style is inflexible which directly abates its possibility of wide industrial applications, such as Search Engine Optimization (SEO). Albeit the test yielded from deep learning-based methods demonstrates a style very close to the one written by a human, its pronounced pitfall that the text context is not under control commonly causes the generated text to be meaningless.

The approach we propose below builds upon the advantages of both and eliminates their respective disadvantage so that the generated text holds human-writing styles whilst the context logic being controllable. The architecture can be seen through Fig. 1.

   \begin{figure}[thpb]
      \centering
      \includegraphics[scale=0.3]{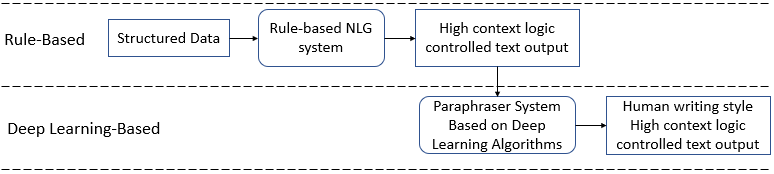}
      \caption{Enhanced Nature Language Generation System Architecture}
      \label{figurelabel}
   \end{figure}

Even though the rule-based and deep-learning-based models work collaboratively when generating text, their model building and training processes are completely different. The rule-based model is constructed through predefined instructions with no training required. To compare it against the deep learning model, the parameters of the deep learning model need to be trained. The dataset utilized to train the deep learning model is not the data derived from the rule-based model but rather the generic data after cleansing and preprocessing to ensure that the content and the paragraphing of which are semantically and grammatically correct and maintain human-writing style so that the trained deep learning model can yield text, satisfying our expectation.

\section{PRELIMINARIES}

This section presents the HMCU analysis model that is adopted to compare and evaluate the performance of various NLG model, along with the brief introduction of the essential concepts regarding rule-based as well as deep learning-based NLG techniques that are conducive to understand our proposed model coherently. 

\subsection{HMCU Analysis Model}

The measurement of the NLG system cannot be conducted quantitively, via using metrics such as Mean Square Error or Mean Absolute Error which are quantitative metrics that reflect the loss between predicted outcomes and true labels of data samples. Therefore, in this paper we propose a novel method that evaluates the NLG system in terms of four predefined metrics. The metrics applied to measure the NLG system can be summarized into four broad categories, i.e. H, M, C, U, the indications of which are listed below:
\begin{itemize}
\item H(Human language): The similarity between the generated text and human writing text.

\item M(Machine style): The similarity between the generated text and machine writing text.

\item C(Controllable context logic): The degree to which the context logic is under control.

\item U(Uncontrollable context logic): The degree to which the content logic is out of control.
\end{itemize}

H and M evaluate the style of the generated text. H stands for the extent to which the texts manifest Human-writing style. Similarly, M represents the extent to which the texts manifest Machine-writing style. C and U evaluate the extent to which the context logic is under control. These four metrics are not designed to be used in a stand-alone but, rather in an aggregated manner. Any NLG model could be measured by utilizing the combination of these metrics, as shown in Table \uppercase\expandafter{\romannumeral1}.

\begin{table}[]
\caption{HMCU evaluation approach}
\label{HMCU evaluation approach}
\begin{tabular}{l|l|l|}
\cline{2-3}
                                                                                                       & \multicolumn{1}{c|}{\cellcolor[HTML]{656565}\textbf{C}}                                                                            & \multicolumn{1}{c|}{\cellcolor[HTML]{656565}\textbf{U}}                                                                               \\ \cline{2-3} 
                                                                                                       & \cellcolor[HTML]{EFEFEF}Controllable context logic                                                                                 & \cellcolor[HTML]{EFEFEF}Uncontrollable context logic                                                                                  \\ \hline
\rowcolor[HTML]{C0C0C0} 
\multicolumn{1}{|c|}{\cellcolor[HTML]{656565}{\color[HTML]{333333} H}}                                 & \multicolumn{1}{c|}{\cellcolor[HTML]{C0C0C0}{\color[HTML]{333333} HC}}                                                             & \multicolumn{1}{c|}{\cellcolor[HTML]{C0C0C0}{\color[HTML]{333333} HU}}                                                                \\ \hline
\multicolumn{1}{|l|}{\cellcolor[HTML]{EFEFEF}\begin{tabular}[c]{@{}l@{}}Human\\ language\end{tabular}} & \begin{tabular}[c]{@{}l@{}}The generated text has a\\ human-writing style and\\ the context logic is\\ controllable\end{tabular}   & \begin{tabular}[c]{@{}l@{}}The generated text has a\\ human-writing style and\\ the context logic is hard\\ to control\end{tabular}   \\ \hline
\rowcolor[HTML]{C0C0C0} 
\multicolumn{1}{|c|}{\cellcolor[HTML]{656565}M}                                                        & \multicolumn{1}{c|}{\cellcolor[HTML]{C0C0C0}MC}                                                                                    & \multicolumn{1}{c|}{\cellcolor[HTML]{C0C0C0}MU}                                                                                       \\ \hline
\multicolumn{1}{|l|}{\cellcolor[HTML]{EFEFEF}\begin{tabular}[c]{@{}l@{}}Machine\\ style\end{tabular}}  & \begin{tabular}[c]{@{}l@{}}The generated text has a\\ machine-writing style and\\ the context logic is\\ controllable\end{tabular} & \begin{tabular}[c]{@{}l@{}}The generated text has a\\ machine-writing style and\\ the context logic is hard\\ to control\end{tabular} \\ \hline
\end{tabular}
\end{table}

Commonly, the text generated from the rule-based NLG model tends to fall into the MC category whereas the one generated from the deep learning-based NLG model is likely to be categorized into HU when the models are constructed and trained appropriately.
 
The ideal output-text of the NLG system should fall in the HC category, which is a rather improbable outcome. Thus, we must balance different constraints out and consider the characteristics and peculiarities of the real-world applications before deploying the model into production. In the empirical research section, the applications of utilizing HMCU approaches to evaluate different NLG systems are demonstrated.

\subsection{Rule-based NLG Techniques}

The incentives for using NLG techniques is to obtain natural language text for a wide spectrum of inputs rather than canned and fixed text with simple templates. Namely, the rule-based NLG techniques generate text by predefined rules. Albeit high quality in terms of human readability is guaranteed, the stiff style is the most salient difficulty to overcome.

Rule-based NLG systems were introduced by Reiter and Dale \cite{c1}. They listed several tasks falling into the problems that the NLG system could resolve, including content determination, discourse planning, sentence aggregation, lexicalization, referring expression generation, and linguistic realization. In addition to these tasks, they also proposed a pipeline with three stages to tackle these tasks just described, which composes of text planning, sentence planning, linguistic realization.

The core of rule-based NLG techniques is the idea of generation as a deterministic decision-making process eliminating some forms of stochasticity. For example, the rule-based grammar credits the capacity to yield previously unseen output through recursive rules\cite{c2}. Although these rules preserve the randomness of generated texts to some extent, the outputs are still deterministic. Given the same inputs, the outputs are always identical under specific rules. 

\subsection{Deep Learning-Based NLG techniques}

Nowadays, to the best of our knowledge, the majority of deep learning techniques that are adopted to implement natural language generation applications fall into two categories, either recurrent neural network(RNN) such as Long Short-Term Memory\cite{c7} or Transformer\cite{c4} models, as listed below:

\begin{itemize}
\item The rapid innovation and development of deep learning techniques enabled their applications on NLG systems. Long Short-Term Memory (LSTM) models and their multiple variations have been extensively applied to Nature Language Processing (NLP) tasks because of their prowess in handling time series data. Recently, studies have shown that LSTMs can be leveraged to generate whole sentences that are analogous to the ones written by a human. For instance, this paper\cite{c6} proposed an encoder-decoder structure complemented by a discriminator that could produce high-readable and context awareness sentences, which is the state-of-the-art approach.

\item In the domain of deep learning, sequence-to-sequence models are commonly using an encoder-decoder architecture to complete tasks like machine translation where one sequence written in one language is converted to a sequence of another language while the meaning is fully preserved, which is considered a similar task to paraphrasing. Transformer models that adopt the encoder-decoder architecture have been applied to a wide range of NLG applications due to its unique architecture and high performance. GPT-3\cite{c3} is one of the remarkable examples of the transformer model being utilized on some NLP tasks and has very recently demonstrated brilliant results. However, its tremendous size becomes the major hindrance that thwarts its popularity.

\end{itemize}

Due to the maturity and reliability of the LSTM model in NLG related tasks, we adopt it as the deep-learning technique to further process the text generated from the rule-based model. The transformed model could also be an option that is used as an alternative to the LSTM model.  

\section{Enhanced natural generation system with hybrid approaches}

Machines can generate text with controllable context logic to accord with the intention of the human through Rule-based NLG techniques; however the generated output probably maintain a similar writing style (machine writing style) since the output text is based on presets templates. On the other hand, machines can generate human writing style text based on deep learning techniques but the context logic of the output cannot be controlled by a human. The purpose of this paper is to propose a novel hybrid method to enhance NLG systems and make them capable of performing with a highly controllable context logic and a flexible human writing style.

   \begin{figure}[thpb]
      \centering
      \includegraphics[scale=0.39]{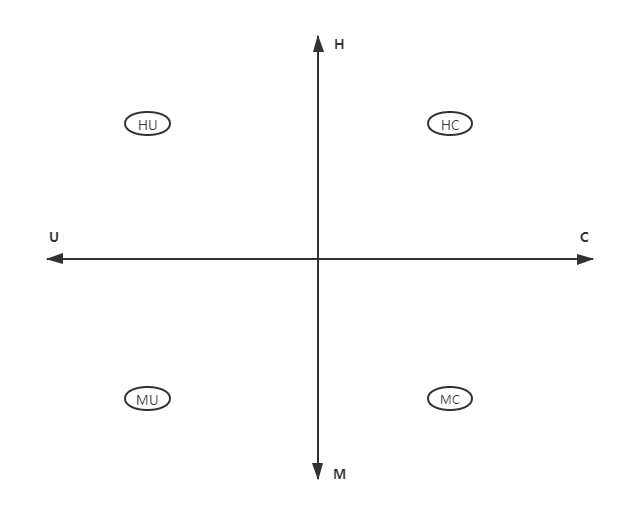}
      \caption{HMCU analysis model}
      \label{figurelabel}
   \end{figure}

Fig. 2 shows our proposed text quality analysis model called HMCU, along with the output quality of the text generated by machines that can be analyzed through this model. Given that HMCU is a text quality evaluation method, it can be used in actual applications independently or combined with other evaluation methods.

The quality of the text generated by machines through rule-based techniques is usually in the fourth quadrant in Fig. 2, and the quality of the text generated based on deep learning techniques usually in the second quadrant in Fig. 2. However, ideally, text quality should be in the first quadrant, i.e. closer to human writing style, with the context logic controllable by humans. Therefore, the main task of the enhanced NLG system is to allow the final output text quality to move to the first quadrant as much as possible.

In general, our proposed NLG system works in the form of a pipeline, as demonstrated in fig. 1. It firstly takes the generated text with high MC from the rule-based model and feeds it to a deep learning model that is able to yield text with high HC. The generated text from the deep learning models has high HU because it is unrestrained from any established rules leading to output styles that are not similar to the preset templates. Therefore the style of generated text from the deep learning models is flexible and closer to the human writing style. 

Although the context logic cannot be controlled by humans if the text is generated based on deep learning models, people can restrict the deep learning part to only paraphrasing the text generated from the rule-based part. This means that in our scheme, the deep learning part is not used for the creation of text, but rather for  paraphrasing of sentences. To make the final context logic controllable, the deep learning part should paraphrase the text sentence by sentence rather than the entire text in one go. It is worth pointing out that the training data of the deep learning model is generic data rather than text data generated based on rules to avoid making the writing style inflexible, which would be the case if people were to use rule-based output as training data. This means that the deep learning model can paraphrase any sentence, and make the style of the paraphrasing sentence closer to the human style while maintaining the original meaning(context logic).

The NLG system we propose can utilize the advantages of both methods while eliminating their disadvantages to some extent. As shown in fig. 1, in this example, our system takes input as structure data that consists of some attributes, like the time the event occurred, the people involved, the location where the event happened, etc. The rule-based system then produces text in accord with these given attributes, which is subsequently used as an input to the deep learning-based system aiming for the final output to both contain  human-writing style and controllable context logic.

\section{EMPIRICAL RESEARCH}

A real-time news generation system will be used to test these ideas in this section, and the output text from rule-based system, deep learning-based system, and a proposed hybrid system will be compared to evaluate the performance of the proposed system.

\subsection{Case Study}

One of the many application domains of the NLG system is the real-time news generation that curtails enormously the time for the news to be written properly\cite{c8}. We have developed 3 systems for news generation using rule-based, deep learning-based, and hybrid-based techniques separately. The output text will be used to evaluate the performance of these systems from two aspects, i.e. controllability of context logic and flexibility of writing style. We assume that a more flexible writing style indicates attributes closer to the human writing style.

To make the experimental results more credible, the testing data has been split into 5 groups. From the perspective of the controllability of context logic, the text generated based on rules has the highest average context logic similarity, which is 0.744. In contrast, the text generated based on LSTM has the lowest average context logic similarity at 0.059. Besides, the average context logic similarity of the text generated by the hybrid method is 0.173 (detailed evaluation methods and steps will be shown in Section F). The experimental results show that the text generated based on rules still has the highest controllable context logic, but the hybrid method can greatly improve the controllability of the context logic of the text generated by the deep learning approaches.

From the perspective of the flexibility of writing style, the text generated using the LSTM technique is the farthest from the machine writing style, with the similarity to the machine writing style being 0.043. On the contrary, the text generated using the rule-based approaches has the highest similarity with a machine writing style, at 0.242. The similarity to a machine writing style of the text generated by the hybrid approach is 0.098, which is much lower than the similarity achieved by the rule-based technique. This means that compared to the rule-based text generation technique, the hybrid approach can greatly improve the flexibility of the writing style of the text.

The hybrid approach is a way to balance the performance of the generated text between the controllability of context logic and flexibility of writing style. It can make the output text not only reach the degree of human acceptance and understanding of the context logic of the text but also keep the output text within a flexible writing style.

\subsection{Rule-based Natural Language Generation system}

We got 900 short news data from a news website(i.e. indiatimes.com) and split it into 2 parts, i.e. 675 rows for training data and 225 rows for testing data. The testing data has been divided into 5 groups, where each group includes 45 rows of testing data. Notice that the training data here is not prepared for the paraphrasing system, but for the system that only uses the LSTM technique for text generation.

Then, we prepared a one-to-one structured data for each row of the testing data. The structure of the structured data  as used in this project has been designed as follows:\\

\begin {math}
\lbrack\lbrace"subject":"","verb":"","object":"","reason":"","purpose":"","area":"","date":"","week":"","year":"","month":""\rbrace,\lbrace"subject":"","verb":"","object":"","reason":"","purpose":"","area":"","date":"","week":"","year":"","month":""\rbrace\rbrack
\\
\end {math}

Each dictionary in the list corresponds to a sentence of news, and each dictionary contains various elements of the news, such as date, subject, and area. This is not the only way to design structured data. In actual projects, people can also design different structured data according to their preferences and needs.

With structured data, people can design detailed rules and templates to complete the entire Rule-based natural language generation system\cite{c9}. For example, the following is an example generated by a Rule-based natural language generation system implemented by us.\\

\textit{“Germany is well placed avoid wave of coronavirus. Moreover, the absence of a vaccine meant social distancing were necessary because Scholz said on Friday. Moreover, we are living with the virus. Besides, it will change we can have new medical therapies. Moreover, we have to organise our lives in order to avoid a second wave.”}\\

Through this example, people can know that the context logic of these texts can be understood by humans, and the context logic of the content can be controlled by the NLG system designer. However, when the number of samples generated becomes large, people will find that the writing style of these texts is very similar (i.e. inflexible) and has a very high machine writing style. Numbers of problems will occur in the actual use scenarios, such as the user's aesthetic fatigue in reading, not conducive to SEO, and so on.

\subsection{Natural Language Generation system based on LSTM}

This part is not needed in the hybrid approach we proposed, but to be able to compare the performance of text generated by various techniques, we developed an NLG system using the LSTM technique.

The LSTM NLG model needs to be trained before using it. The training data used in this for this part is one described in subsection B. After training the model, one can enter keywords into the model, and the system will automatically create sentences or news based on the trained LSTM model.

To evaluate the model developed, we extracted key information from each row of the testing data and entered it into the LSTM NLG system. After that, the system automatically generated different pieces of news. One of the news samples generated by the NLG system based on LSTM is as follows:\\

\textit{“Germany is well to the bodies. The absence of the virus to the bodies. We are living to function. It will change to Medical to proper to proper to proper to proper to proper to proper to proper to proper to proper to proper to proper to proper to proper to proper to proper. We have to the country.”}\\

It can be seen from the news sample that although this news looks very flexible in writing style the context logic of the content is uncontrollable. Thus, this news has no real meaning. Many people try to use the LSTM technique to generate fake news, but these fake news cannot be applied in the actual environment, because in practice real news are written based on factual information.

\subsection{Paraphrasing system based on Deep Learning techniques}

The paraphrasing system can be realized based on deep learning techniques. In this paper we used the LSTM technique as the paraphrasing system\cite{c10}. Notice that the paraphrasing system and the system presented in subsection C are not the same systems, despite both being based on LSTMs.

The Sequence-to-Sequence model will be used in this section. The purpose of the paraphrasing part is to allow the model to automatically paraphrase arbitrary sentences, rather than just paraphrasing news-related sentences\cite{c11}. Therefore, the training data of the paraphrasing system is not the specific news data mentioned in subsection B, but generic data.

The data set, called ParaNMT-50M, will be used as the training data for the paraphrasing system in this paper\cite{c12}. However, people can choose other data-sets as the training data for the paraphrasing section according to their preferences.

After the paraphrasing system has been developed and trained, any input sentence can be paraphrased. For example, when entering "how are you", the paraphrasing system can paraphrase the sentence to “how are you doing?”, “how do you feel?” and so on.
 
\subsection{Integrated NLG system}

The integrated system will incorporate the systems presented in subsections B and D. Specifically, we first design the structured data of news and then we add them as input into the system of subsection B. Following this, the system of subsection B will produce an output text, which will be entered into the system of subsection D. After that, the system of subsection D will produce an output text, which will then be grammatically checked to correct any grammatical errors generated by the paraphrasing part, thereby producing the final output result.

The integrated NLG system combines a rule and deep learning-based techniques, which can generate text with controllable context logic and a flexible writing style. For example, the following sentence is an example output as generated by the proposed system:\\

\textit{“Germany is well placed away from the virus. As if they were not given any possibility of having a vaccine, did it require to be a social denial because Scholz said. We live with the virus. Besides, it will change a new medical treatment. In addition, we have to organize our lives to avoid the second wave.”}\\

From this output text, it can be seen that the context logic of the text is roughly consistent with the real news events. On the premise of ensuring that the context logic is controllable, the writing style of news texts becomes very flexible. However, it is worth noting that there are still some sentences whose context logic deviates from the original news events, e.g. the system paraphrased the sentence "Moreover, the absence of a vaccine meant social distancing was necessary because Scholz said on Friday." to the sentence “As if they were not given any possibility of having a vaccine, did it require to be a social denial because Scholz said. ”, which changes a bit the original context logic. Therefore, we need to further study how to solve or avoid such small deviations in the future.

\subsection{Discussion of Results and Evaluation}

The performance of the text will be evaluated from two aspects, controllability of context logic and flexibility of writing style.

Controllability of context logic can be quantified via calculating the similarity between the generated text and real news. Firstly, we set the text generated based on rules, the text generated based on LSTM and the text generated by the proposed hybrid approach as the corpus, and then used the real news corresponding to the text as the input, in order to calculate the similarity between them (we named this similarity as contextual logic similarity). The contextual logic similarity of each news(testing data) can be calculated first, and then the average contextual logic similarity of each group can be calculated. Then, the contextual logic similarity of these 5 groups can be averaged again to obtain the final average contextual logic similarity.

\begin{table}[]
\caption{Contextual logic similarity}
\begin{tabular}{|l|l|l|l|}
\hline
\rowcolor[HTML]{C0C0C0} 
\multicolumn{1}{|c|}{\cellcolor[HTML]{C0C0C0}Group} & \begin{tabular}[c]{@{}l@{}}The contextual \\ logic similarity\\ of Rule-based\\ techniques\end{tabular} & \begin{tabular}[c]{@{}l@{}}The contextual\\ logic similarity\\ of LSTM\\ techniques\end{tabular} & \begin{tabular}[c]{@{}l@{}}The contextual\\ logic similarity\\ of the\\ Hybrid approach\end{tabular} \\ \hline
1                                                   & 0.553                                                                                                   & 0.005                                                                                            & 0.15                                                                                                 \\ \hline
2                                                   & 0.709                                                                                                   & 0.111                                                                                            & 0.182                                                                                                \\ \hline
3                                                   & 0.893                                                                                                   & 0.012                                                                                            & 0.167                                                                                                \\ \hline
4                                                   & 0.835                                                                                                   & 0.029                                                                                            & 0.231                                                                                                \\ \hline
5                                                   & 0.728                                                                                                   & 0.138                                                                                            & 0.136                                                                                                \\ \hline
\textbf{average}                                    & \textbf{0.744}                                                                                          & \textbf{0.059}                                                                                   & \textbf{0.173}                                                                                       \\ \hline
\end{tabular}
\end{table}

Analyzing the results presented in Table \uppercase\expandafter{\romannumeral2}, one can see that the text generated based on Rule-based techniques has a very high contextual logic similarity, with a similar trend observed across all groups. The contextual logical similarity of the text generated based on the LSTM technique is very low and unstable, e.g. the values of group 1 and group 3 are very low, which are 0.005 and 0.012 respectively. This means that it is likely that the text generated by these two groups is completely inconsistent with the actual news event. The contextual logic similarity between the text generated by the hybrid approach and the real news is relatively stable and is between 0.13 and 0.25.

Therefore, the contextual logic similarity of the text generated based on the rule is the highest, followed by the text generated by a hybrid approach. Through actual case studies, we can find that the context logic of text generated based on the hybrid approach is understandable, though there are a few sentences that still have problems. The contextual logic similarity of the text generated based on LSTM is extremely low, which does not coincide with the original news event.

On the other hand, the flexibility of writing style can be quantified by comparing the similarity between the text generated by each approach. The higher the similarity between the texts generated based on the same approach, the less flexible the writing style of the text will be, that is, the generated text will be closer to the machine writing style. This similarity can be called machine writing style similarity.

We calculated the similarity for 5 groups of the testing data separately. In each group, people can arbitrarily select a generated text as a benchmark. Hereafter, the similarity between the other generated texts and the benchmark can be calculated, and then the average machine writing style similarity within each group can be calculated. Finally, people can calculate the average machine writing style similarity between these 5 groups.

\begin{table}[]
\caption{Machine writing style similarity}
\begin{tabular}{|l|l|l|l|}
\hline
\rowcolor[HTML]{C0C0C0} 
\multicolumn{1}{|c|}{\cellcolor[HTML]{C0C0C0}Group} & \begin{tabular}[c]{@{}l@{}}Machine writing\\ style similarity\\ of Rule-based\\ techniques\end{tabular} & \begin{tabular}[c]{@{}l@{}}Machine writing\\ style similarity\\ of LSTM\\ techniques\end{tabular} & \begin{tabular}[c]{@{}l@{}}Machine writing\\ style similarity\\ of the\\ Hybrid approach\end{tabular} \\ \hline
1                                                   & 0.254                                                                                                   & 0.037                                                                                             & 0.108                                                                                                 \\ \hline
2                                                   & 0.247                                                                                                   & 0.037                                                                                             & 0.103                                                                                                 \\ \hline
3                                                   & 0.226                                                                                                   & 0.032                                                                                             & 0.09                                                                                                  \\ \hline
4                                                   & 0.243                                                                                                   & 0.082                                                                                             & 0.087                                                                                                 \\ \hline
5                                                   & 0.238                                                                                                   & 0.025                                                                                             & 0.104                                                                                                 \\ \hline
\textbf{average}                                    & \textbf{0.242}                                                                                          & \textbf{0.043}                                                                                    & \textbf{0.098}                                                                                        \\ \hline
\end{tabular}
\end{table}

Table \uppercase\expandafter{\romannumeral3} shows the machine writing style similarity of the text generated by each group based on different approaches. The text generated via the rule-based technique demonstrated  the highest machine writing style similarity, which indicates that the writing style of the text generated via the rule-based technique is inflexible. The proposed hybrid approach can increase the flexibility of the generated text, as evidenced through the fact that the the machine writing style similarity dropped from 0.242 (rule-based techniques) to 0.098 (hybrid approach). Although the LSTM approach has the lowest machine writing style similarity, its context logic is uncontrollable and hard to apply in actual environments currently.

Therefore, the hybrid approach based on rule and deep learning-based techniques we proposed in this paper can make the Natural Language Generation system achieve a balance between the controllability of context logic and flexibility of writing style. The hybrid approach can make the generated text have controllable context logic, along with a flexible writing style.

\section{Conclusion \& FUTURE WORK}

Although using a hybrid approach can make most of the context logic of the resulting text controllable, there are still some sentences where the context logic cannot be controlled. We will do more studies for this research challenge in the future.

Besides, we put forward an evaluation approach which called HMCU in this paper, but the quantitative indicators and calculation methods are still not mature enough. We will design more mature quantitative indicators and calculation methods in the future.

\section{Acknowledge }

We are very grateful for the very useful suggestions and guidance that Professor Ehud Reiter offered throughout the development and writing phases of this paper.

\bibliographystyle{ieeetr}
\bibliography{main}
\end{document}